\documentclass[10pt,onecolumn,letterpaper]{article}

\usepackage{iccv}
\usepackage{times}
\usepackage{epsfig}
\usepackage{graphicx}
\usepackage{amsmath}
\usepackage{amssymb}
\usepackage{amsbsy}
\usepackage{color}
\usepackage{url}

\newcommand{\SKIP}[1]{} 

\newcommand{\mbegin} {\left [ \begin{array}}

\newcommand{\mend}   {\end{array} \right ]}

\newcommand{\detbegin} {\left | \begin{array}}
\newcommand{\detend}   {\end{array} \right |}

\newcommand{\vbegin} {\left ( \begin{array}{c}}

\newcommand{\vend} {\end{array}\right )}

\def\squareforqed{\hbox{\rlap{$\sqcap$}$\sqcup$}}

\def\qed{\ifmmode\squareforqed\else{\unskip\nobreak\hfil
	\penalty50\hskip1em\null\nobreak\hfil\squareforqed
	\parfillskip=0pt\finalhyphendemerits=0\endgraf}\fi}

\newcommand{\showeqnlabel}{
	\hbox to 0pt{\quad\quad\relax\fbox{\scriptsize\rm\eqnlblx}%
	\gdef\eqnlblx{xxxx}}} \newcommand{\eqnlblx}{}

\def\@eqnnum{\rm (\theequation)\showeqnlabel}

\newcommand{\nofig}[1]{\centerline{\bf Figure here}}

\newcommand*\Ell{\ensuremath{\boldsymbol\ell}}

\usepackage[pagebackref=true,breaklinks=true,letterpaper=true,colorlinks,bookmarks=false]{hyperref}

\iccvfinalcopy 

\ificcvfinal\fi
\begin{document}

\title{``Maximizing Rigidity" Revisited: a Convex Programming Approach for Generic 3D Shape Reconstruction from Multiple Perspective Views} 

\author{Pan Ji\\
University of Adelaide\\
{\tt\small pan.ji@adelaide.edu.au}
\and
Hongdong Li, Yuchao Dai\\
Australian National University\\
{\tt\small firstname.lastname@anu.edu.au}
\and
Ian Reid\\
University of Adelaide\\
{\tt\small ian.reid@adelaide.edu.au}
}

\maketitle

\begin{abstract}
Rigid structure-from-motion (RSfM) and non-rigid structure-from-motion (NRSfM) have long been treated in the literature as separate (different) problems.  Inspired by a previous work which solved directly for 3D scene structure by factoring the relative camera poses out, we revisit the principle of ``maximizing rigidity'' in structure-from-motion literature, and develop a unified theory which is applicable to both rigid and non-rigid structure reconstruction in a rigidity-agnostic way.  We formulate these problems as a convex semi-definite program, imposing constraints that seek to apply the principle of minimizing non-rigidity. Our results demonstrate the efficacy of the approach, with state-of-the-art accuracy on various 3D reconstruction problems.\footnote{Source codes will be available at \url{https://sites.google.com/site/peterji1990/resources/software}.}
\end{abstract}
\section{Introduction}
Structure-from-motion (SfM) is the problem of recovering the 3D structure of a scene from multiple images taken by a camera at different viewpoints. When the scene structure is rigid the problem is generally well defined and has been much studied \cite{ullman1979interpretation,longuet1987computer,tomasi1992shape,hartley2003multiple}, with rigidity at the heart of almost all vision-based 3D reconstruction theories and methods. When the scene structure is non-rigid (deforming surface, articulated motion, and \etc), the problem is under-constrained, and constraints such as low dimensionality~\cite{dai2014simple} or local rigidity~\cite{Locally-Rigid:CVPR-2010} have been exploited to limit the set of solutions. 

Currently, non-rigid structure from motion (NRSfM) lags far behind its rigid counterpart, and is often treated entirely separately from rigid SfM.  Part of the reason for this separate treatment lies in the usual formulation of the SfM problem, which approaches the task in two stages: first the relative camera motions \wrt the scene are estimated; then the 3D structure is computed afterwards. In each stage, different methods and implementations have to be developed for rigid and non-rigid scenarios separately because of the different structure priors that are exploited. This has a further disadvantage in that it can be difficult to determine {\em a priori} whether the scene is rigid or nonrigid (and if the latter, in what way).

Therefore, it is highly desirable to have a generic SfM framework that can deal with both rigid and non-rigid motion, which leads to the main theme of this paper. In fact, as early as in the year 1983, Ullman~\cite{ullman1984maximizing} proposed a ``maximizing rigidity'' principle that relies on a non-convex rigidity measure to reconstruct 3D structure from both rigid and non-rigid (rubbery) motion. This idea has resurfaced in various work under the ARAP (``as rigid as possible'') moniker~\cite{parashar2015rigid,sorkine2007rigid}. However, it has not been further developed under the modern view of 3D reconstruction, mainly due to the difficulty in its optimization.
In this paper we revisit Ullman's ``maximizing rigidity'' principle and propose a novel convex rigidity measure that can be incorporated into a modern SfM framework for both rigid and non-rigid shape reconstruction. 

Our proposed formulation yields reconstructions that are more accurate than current state-of-the-art for non-rigid shape reconstruction, and which enforce rigidity when this is present in the scene. This is because our framework aims at maximizing the rigidity while still satisfying the image measurements. We thus achieve a unified theory and paradigm for 3D vision reconstruction tasks for both rigid and non-rigid surfaces. Our method does not need to specify which case (out of the above scenarios) is the target to be reconstructed. The method will automatically output the optimal solution that best explains the observations.



\section{Related Work}
Traditionally, under a perspective camera, the pipeline of RSfM consists of two steps, \ie, a camera motion estimation step and a following structure computation step~\cite{longuet1987computer,hartley1997triangulation,hartley1997defense,hartley2003multiple}; or the camera motion and 3D structure can also be estimated simultaneously through measurement matrix factorization~\cite{tomasi1992shape,sturm1996factorization}. In RSfM literature, most related to our work was by Li \cite{li2010multi}, who proposed an unusual approach to handle SfM which bypasses the motion-estimation step. This method does not require any explicit motion estimation and was called the ``structure-without-motion'' method. 

In contrast to RSfM, NRSfM remains an open active research topic~\cite{dai2014simple,chhatkuli2016inextensible,wang2016template} in computer vision. One of the commonly used constraints in NRSfM is the local rigidity constraint~\cite{Locally-Rigid:CVPR-2010,Salzmann-Template-Free:ICCV-2009}, or in some literature the inextensibility constraint~\cite{Soft-Inextensibility-Template-Free:ECCV-2012}.  

Taylor \etal~\cite{Locally-Rigid:CVPR-2010} formulated a NRSfM framework in terms of a set of linear length recovery equations using local three-point triangles under an orthographic camera, and grouped these ``loosely coupled'' rigid triangles into non-rigid bodies. Varol \etal \cite{Salzmann-Template-Free:ICCV-2009} estimated homographies between corresponding local planar patches in both images. These yield approximate 3D reconstructions of points within each patch up to a scale factor, where the relative scales are resolved by considering the consistency in the overlapping patches. Both methods form part of a recent trend of piece-wise reconstruction methods in NRSfM. Instead of relying on a single model for the full surface, these approaches model small patches of the surface independently.  Vicente and Agapito \cite{Soft-Inextensibility-Template-Free:ECCV-2012} exploited a soft inextensibility prior for template-free non-rigid reconstruction. They formulated an energy function that incorporates the inextensibility prior and minimized it via the QPBO algorithm. Very recently, Chhatkuli \etal \cite{chhatkuli2016inextensible} presented a global and convex formulation for template-less 3D reconstruction of a deforming object by using perspective cameras, where the 3D reconstruction problem is recast as a Second-Order Cone Programming (SOCP) using the Maximum-Depth Heuristic (MDH)~\cite{Perriollat:BMVC08,Perriollat_IJCV_2010,salzmann2011linear}. 

The literature of RSfM and NRSfM advance in relatively independent directions. To the best of our knowledge, the first attempt to unify the two fields was by Ullman~\cite{ullman1984maximizing}, who proposed to use the principle of ``maximizing rigidity' to recover 3D structure from both rigid and non-rigid motion. Ullman's original formulation maintained and updated an internal rigid model of the scene across a temporal sequence. A rigid metric was defined in terms of point distance to measure the deviation from the estimated structure to the internal model. The 3D structure was recovered by minimizing the overall deviation from rigidity (internal model) to a local optimum via a gradient method. Compared to Ullman's method, our method unifies rigid and non-rigid SfM within a convex program, from which we obtain a global optimal solution.



\section{Maximizing Rigidity Revisited}

In this section, we discuss Ullman's maximizing rigidity principle in more detail. Ullman assumed that there is an internal model of the scene and the internal model should be as rigid as possible~\cite{ullman1984maximizing}. Let $\bar{d}_{ij}$ be the Euclidean distance between points $i$ and $j$ of the internal model, and ${d}_{ij}$ the Euclidean distance between points $i$ and $j$ of the estimated structure. A measure of the difference between $\bar{d}_{ij}$ and ${d}_{ij}$ was defined as
\begin{equation}
\Delta_{ij} = \frac{(\bar{d}_{ij} - {d}_{ij})^2}{\bar{d}_{ij}^3}\;.
\end{equation}
Under an orthographic camera model, the pairwise distance ${d}_{ij}$ was directly parameterized as $(x_i - x_j)^2 + (y_i - y_j)^2 + (z_i - z_j)^2$, where $\{(x_i, y_i)\}$ are the known image (coordinate) measurements and $\{z_i\}$ are the unknown depths. Unfortunately, no principled way was provided to handle the general perspective camera model. 
Intuitively, this measure is the least square difference reweighted by the inter-point distance of the internal model. The reweighting indicates that a point is more likely to move rigidly with its nearest neighbors~\cite{ullman1984maximizing}. However, the reweighting also makes Ullman's rigidity measure non-convex in terms of $\bar{d}_{ij}$.

\begin{figure}[!t]
\centering
\includegraphics[width=0.50\linewidth]{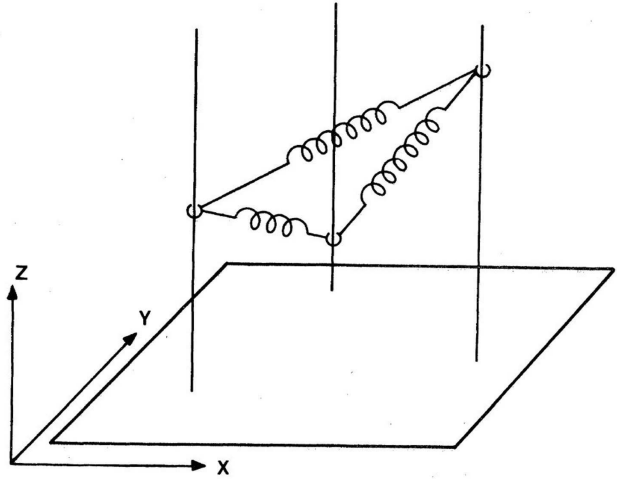}
\caption{A spring model illustration for Ullman's maximizing rigidity principle. Each of the viewed points (three in this example) is constrained to move along one of the rigid rods, and its position along the rod represents its depth. The connecting springs represent the distances between points in the current internal model. The points would slide along the rods until a minimum energy configuration is reached. The final configuration represents the modified internal model. Image and caption are modified from Figure 2 of~\cite{ullman1984maximizing}.}
\vspace{-0.3cm}
\label{fig:ullmanspringmodel}
\end{figure}

Then the problem of determining the most rigid structure can be formulated as minimizing the overall deviation from rigidity $\sum_{ij} \Delta_{ij}$. Since the internal model of the scene is often unknown, the remedy that Ullman proposed was to start from a flat plane (as the initial internal model) and incrementally estimate the internal model and 3D structure. This internal model was shown to converge to a local optimum for both rigid and non-rigid motions~\cite{ullman1984maximizing,grzywacz1987incremental}. See Figure~\ref{fig:ullmanspringmodel} for a spring model illustration of Ullman's method.

From the analysis above, we can identify three major drawbacks of Ullman's method: ({\bf i}) it cannot handle the perspective camera in a principled way; ({\bf ii}) the rigidity measure used is non-convex, which leads to local optimum; ({\bf iii}) it relies on building fully connected graphs (for every pair of points), which, in practice, is often redundant and unnecessary. In the following section, we'll show how these drawbacks can be circumvented by introducing a novel convex rigidity measure which can be further incorporated into an edge-based 3D reconstruction framework for perspective projection. 

\section{Our 3D Shape Reconstruction Model}

In this section, we present our unified model for both rigid and nonrigid 3D shape reconstruction. The core component of our model lies in a novel convex rigidity measure as introduced below. For notation, points are indexed with a subscript $i \in \{1,\cdots,n\}$, and image views (or frames) are indexed with a superscript $k \in \{1,\cdots,m\}$. We assume that the world frame is centered at the camera center and aligned with camera coordinate system.

\subsection{Our Rigidity Measure} 

Ullman's rigidity measure requires to build a fully connected graph within each time frame and penalize distant edges (as distant point pairs are more likely to move non-rigidly). Instead of using a fully connected graph, we build a K-nearest-neighbor graph (K-NNG), which connects each point $i$ to a set of its $K$ nearest neighbors, denoted as $\mathcal{N}(i)$, based on the Euclidean distance on 2D images~\cite{chhatkuli2016inextensible}. We also use a different internal model than Ullman's. Specifically, we define a rigid internal model with inter-point distance $g_{ij} = \max_k \{d_{ij}^k\}_{k = 1,\cdots, m}$, \ie, $g_{ij}$ is the maximal distance between points ${\bf Q}_i$ and ${\bf Q}_j$ over all frames. For rigid shapes, $g_{ij}$ corresponds to the Euclidean distance between ${\bf Q}_i$ and ${\bf Q}_j$, which is invariant over all frames; for non-rigid inextensible shapes, $g_{ij}$ corresponds to the maximal Euclidean distance between ${\bf Q}_i$ and ${\bf Q}_j$ over all frames, which generally equals the Geodesic distance between ${\bf Q}_i$ and ${\bf Q}_j$. For example, for a non-rigidly deforming paper, its internal model corresponds to the flat paper. To enforce rigidity, we define a measure of the total difference between $g_{ij}$ and $d_{ij}^k$ for all $(i,j,k)$ as
\begin{equation}
\label{eq:our-rigidity}
\Delta' = \sum_{i,j\in\mathcal{N}(i),k} |g_{ij}-d_{ij}^k|\;.
\end{equation}
Compared to Ullman'd rigidity measure, our rigidity measure has three merits: ({\bf i}) we significantly reduce the number of edges by using a K-NNG instead of a fully connected graph; ({\bf ii}) our measure is convex, which is crucial for optimization; ({\bf iii}) we use a robust L-1 norm instead of the L-2 norm in the rigidity measures. To make the reconstructed scene as rigid as possible, we need to minimize $\Delta'$. As we will see in the following subsections, our rigidity measure can be naturally incorporated into an edge-based 3D reconstruction framework under perspective projection.

\subsection{Edge-Based Reconstruction}

Given a set of $n$ 2D point correspondences across $m$ images $\{{\bf q}_i^k\}$, our target is to find their 3D coordinates ${\bf Q}_i^k$ in the same global coordinate system. We denote the edge (distance) between the camera center ${\bf O}$ and ${\bf Q}_i^k$, which we call a ``leg'', as ${\ell}_i^k$. Define the angle between legs ${\ell}_i^k$ and ${\ell}_j^k$ as $\theta_{ij}^k$. Clearly, we have $\theta_{ij}^k = \theta_{ji}^k$. We assume that the camera is intrinsically calibrated, so the angles $\theta_{ij}^k$ can be trivially computed. Denote the Euclidean distance between two points ${\bf Q}_i^k$ and ${\bf Q}_j^k$ in the $k^{\rm th}$ frame as $d_{ij}^k$. For rigid motion, $d_{ij}^k$ is constant over frames for the same pair of point correspondences. In the case of non-rigid motions, $d_{ij}^k$ may change over frames, but is bounded by a maximal value $g_{ij}$, \ie, $g_{ij} = \max_k \{d_{ij}^k\}_{k = 1,\cdots, m}$.

\begin{figure}[!t]
\centering
\includegraphics[width=0.95\linewidth]{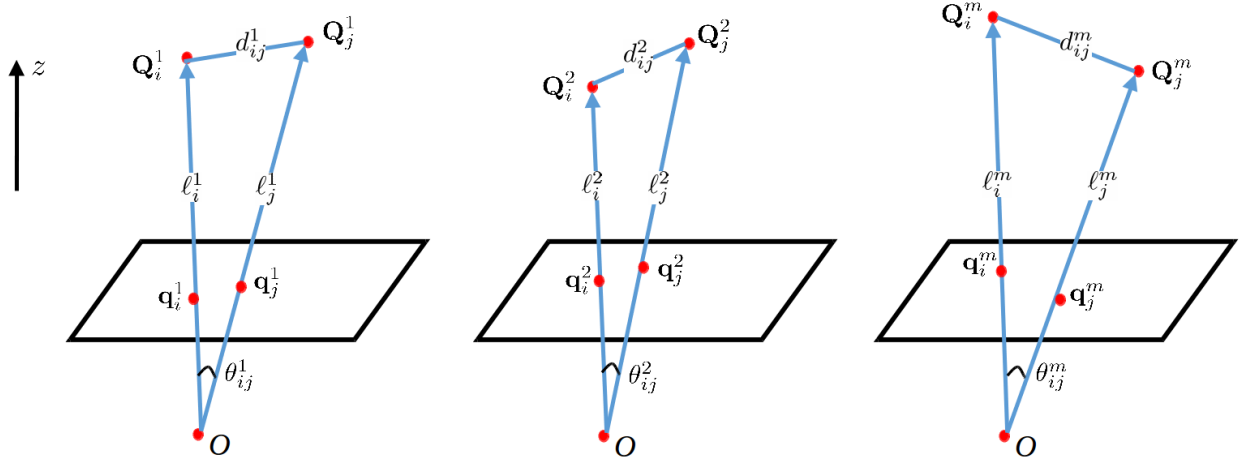}
\caption{Viewing triangles of a pair of 3D points ${\bf Q}_i$ and ${\bf Q}_j$ in different views. We use {O} to denote the camera center from which we draw viewing ray {O}${\bf Q}_i^k$ intersecting with the image plane at ${\bf q}_i^k$. For example, $\bigtriangleup${O}${\bf Q}_i^1{\bf Q}_j^1$ forms a viewing triangle in the first view for ${\bf Q}_i$ and ${\bf Q}_j$.}
\vspace{-0.3cm}
\label{fig2:viewingtriangle}
\end{figure}

Motivated by \cite{li2010multi}, we build our model based on viewing triangles formed by each pair of points to compute the 3D structure. See Figure~\ref{fig2:viewingtriangle} for an illustration. Note that the viewing triangles can only be formed with points of the same frame.

Within each viewing triangle, we have a basic equation following the cosine law
\begin{equation}
{\ell_i^k}^2 + {\ell_j^k}^2 - 2 \ell_i^k\ell_j^k\cos{\theta_{ij}^k} = {d_{ij}^k}^2\;.
\end{equation}

We can rewrite this equation in a matrix form as
\begin{equation}
\label{eq2:coslaw}
\begin{bmatrix} \ell_i^k & \ell_j^k \end{bmatrix}
\begin{bmatrix} 1 & -\cos{\theta_{ij}^k} \\ -\cos{\theta_{ij}^k} & 1 \end{bmatrix}
\begin{bmatrix} \ell_i^k \\ \ell_j^k \end{bmatrix} = {d_{ij}^k}^2\;.
\end{equation}
With all viewing triangles, we can construct a system of quadratic equations of the above form in terms of the unknowns $\ell_i^k$, $\ell_i^k$ and $d_{ij}^k$.

Stack all the legs $\ell_i^k$ into a vector $\Ell = [{\Ell^1}^T \; \cdots \; {\Ell^m}^T]^T$, with the leg vector for the $k^{\rm th}$ frame $\Ell^k = [\ell_1^k\;\cdots\;\ell_n^k]^T$. Define the cosine-matrix as ${\bf C}^k\in\mathbb{R}^{n\times n}$ with the diagonal elements as one and off-diagonal elements as $c_{ij}^k = -\cos{\theta_{ij}^k}$. Let ${\bf e}_i$ be the unit basis vector (\ie, all 0 but 1 at the $i^{\rm th}$ entry), and define the diagonal matrix ${\bf E}_{ij} = {\rm diag}({\bf e}_i+{\bf e}_j)$. Eq.~\eqref{eq2:coslaw} can then be rewritten as
\begin{equation}
{\Ell^k}^T {\bf E}_{ij}^T {\bf C}^k {\bf E}_{ij}\Ell^k = {d_{ij}^k}^2\;.
\end{equation}
Note that the matrix ${\bf A}_{ij}^k \doteq {\bf E}_{ij}^T {\bf C}^k {\bf E}_{ij}$ is highly sparse with only four non-zero elements.

The edge-based 3D reconstruction problem becomes
\begin{subequations}
\begin{align}
               & {\rm Find}\;\; \Ell, {\bf d} \\
               \label{eq:homogeneous}
{\rm s.t.}\;\; & {\Ell^k}^T {\bf A}_{ij}^k \Ell^k = {d_{ij}^k}^2\;,\\
               & \ell_i^k \geq 0, \;d_{ij}^k \geq 0,\;\forall (i,j,k)\;,
\end{align}
\end{subequations}
where ${\bf d}$ is a vector containing all $d_{ij}^k$.

However, the above formulation is not well constrained because: (i) the scale of the solutions cannot be uniquely determined due to the homogeneous equation in~\eqref{eq:homogeneous}; (ii) a trivial all-zero solution always exists; (iii) for non-rigid motions, we only have one equality constraint for each $d_{ij}^k$, which is insufficient to get deterministic solutions\footnote{For rigid motions, $d_{ij}^{k} = d_{ij}$, $\forall k\in\{1,\cdots,m\}$, and we have sufficient constraints for $d_{ij}$. But in many cases, we don't know whether the scene is rigid or non-rigid {\it a priori}.}. The scale ambiguity is intrinsic to 3D reconstruction under the perspective camera model~\cite{hartley2003multiple}. In practice, we can fix a global scale of the scene by normalizing $\Ell$ or ${\bf d}$. 

\paragraph{Maximum-Leg Heuristic (MLH).}

To get the desired solutions, we apply a so-called Maximum-Leg Heuristic (MLH). After fixing the global scale, we want to maximize the sum of all legs $\ell_i^k$ under the non-negative constraint, or equivalently
\begin{subequations}
\label{eq:initformulation}
\begin{align}
     \min_{\Ell,{\bf d}}\;\; & -\sum\limits_{i,k} \ell_i^k \\               
{\rm s.t.}\;\; & {\Ell^k}^T {\bf A}_{ij}^k \Ell^k = {d_{ij}^k}^2\;,\\
               & \ell_i^k \geq 0, \;d_{ij}^k \geq 0,\;\forall (i,j,k)\;.
\end{align}
\end{subequations}
In this way, trivial solutions are avoided. Note that, in the NRSfM literature, there is a commonly used heuristic called Maximum Depth Heuristic (MDH)~\cite{Perriollat:BMVC08,Perriollat_IJCV_2010,salzmann2011linear}, which maximizes the sum of all depths under the condition that each depth and distance are positive. Our MLH virtually plays the same role as MDH because under a perspective camera, we have
$\ell_i^k = z_i^k\|\hat{\bf q}_i^k\|_2$, where $z_i^k$ represents the depth of the $i^{\rm th}$ point in the $k^{\rm th}$ frame (\ie, ${\bf Q}_i^k$), and $\hat{\bf q}_i^k = {\bf K}^{-1}[{{\bf q}_i^k}^T \; 1]^T$.

\section{Convex Program for 3D Shape Reconstruction}
Incorporating our rigidity measure in~\eqref{eq:our-rigidity} into~\eqref{eq:initformulation}, we get our overall formulation for 3D shape reconstruction as follows
\begin{subequations}
\begin{align}
     \min_{\Ell,{\bf d}, {\bf g}}\;\; & -\sum_{i,k} \ell_i^k + \lambda \sum_{i,j,k}|g_{ij}-d_{ij}^k| \\  
               \label{eq:formulation-quad}
{\rm s.t.}\;\; & {\Ell^k}^T {\bf A}_{ij}^k \Ell^k = {d_{ij}^k}^2\;,\\
               & d_{ij}^k \leq g_{ij}\;, \\
               \label{eq:formulation-gij}
               & \sum_{ij} g_{ij}^2 = 1\;,\\
               & \ell_i^k \geq 0, \;d_{ij}^k \geq 0\;,
               \forall \big(i,j\in\mathcal{N}(i),k\big)\;,
\end{align}
\end{subequations}
where $\lambda > 0$ is a trade-off parameter, and the equality constraint~\eqref{eq:formulation-gij} fixes the global scale of the reconstructed shape. However, the above formulation is still non-convex due to the quadratic terms in the both sides of Eq.~\eqref{eq:formulation-quad} and in the left-hand-side of Eq.~\eqref{eq:formulation-gij}. To make it convex, we first define $\hat{g}_{ij} = {g_{ij}}^2$ and $\hat{d}_{ij}^k = {d_{ij}^k}^2$. We then change our formulation to the following form
\begin{subequations}
\begin{align}
     \min_{\Ell,\hat{\bf d}, \hat{\bf g}}\;\; & -\sum_{i,k} \ell_i^k + \lambda \sum_{i,j,k}(\hat{g}_{ij}-\hat{d}_{ij}^k) \\  
               \label{eq:formulation-quad2}
{\rm s.t.}\;\; & {\Ell^k}^T {\bf A}_{ij}^k \Ell^k = \hat{d}_{ij}^k\;,\\
               \label{ineq:dg}
               & \hat{d}_{ij}^k \leq \hat{g}_{ij}\;, \\
               \label{eq:formulation-gij2}
               & \sum_{ij} \hat{g}_{ij} = 1\;,\\
               & \ell_i^k \geq 0, \;\hat{d}_{ij}^k \geq 0\;,
               \forall \big(i,j\in\mathcal{N}(i),k\big)\;,
\end{align}
\end{subequations}
where we approximate $|{g}_{ij}-{d}_{ij}^k|$ with $|\hat{g}_{ij}-\hat{d}_{ij}^k|$, and drop the absolute value operator as we have an inequality constraint~\eqref{ineq:dg} to make sure $\hat{g}_{ij}-\hat{d}_{ij}^k$ is always non-negative.
Due to~\eqref{eq:formulation-quad2}, our formulation turns out to be a quadratically constrained quadratic program (QCQP), which is unfortunately still a non-convex and even NP-hard problem for indefinite ${\bf A}_{ij}^k$~\cite{pardalos1991quadratic,sahni1974computationally}.

\subsection{Semi-Definite Programming (SDP) Relaxation}

We now show how our formulation can be converted to a convex program using SDP relaxation. Note that we have ${\Ell^k}^T {\bf A}_{ij}^k \Ell^k = {\rm tr}({\bf A}_{ij}^k\Ell^k{\Ell^k}^T)$.
We can introduce auxiliary variables ${\bf Y}^k = \Ell^k{\Ell^k}^T$ for $k = 1, \cdots ,m$. Then Eq.~\eqref{eq:formulation-quad2} equivalently becomes two equality constraints
${\rm tr}({\bf A}_{ij}^k {\bf Y}^k) =  \hat{d}_{ij}^k\;,\;
{\bf Y}^k                          =  \Ell^k{\Ell^k}^T$.
We can directly relax the last non-convex equality constraint ${\bf Y}^k = \Ell^k{\Ell^k}^T$ into a convex positive semi-definiteness constraint ${\bf Y}^k \succeq \Ell^k{\Ell^k}^T$~\cite{d2003relaxations}. Using a Schur complement, ${\bf Y}^k \succeq \Ell^k{\Ell^k}^T$ can be reformulated~\cite{boyd2004convex} as
$\begin{bmatrix} 1 & {\Ell^k}^T \\ {\Ell^k} & {\bf Y}^k \end{bmatrix} \succeq 0$.
Ideally, ${\bf Y}^k$ should be a rank-one matrix. But, after the relaxation, the rank constraint for ${\bf Y}^k$ may not be maintained. We can minimize ${\rm tr}({\bf Y}^k)$ as a convex surrogate of ${\rm rank}({\bf Y}^k)$.\footnote{For positive semi-definite ${\bf Y}^k$, ${\rm tr}({\bf Y}^k) = \|{\bf Y}^k\|_*$, and the nuclear norm is a well-known convex surrogate for the rank.}

Our formulation becomes an SDP written as:
\begin{subequations}
\begin{align}
     \label{eq:obj}
     \min_{\Ell,\hat{\bf d}, \hat{\bf g}, {\bf Y}^k}\;\; & \sum_k {\rm tr}({\bf Y}^k)-\lambda_1 {\bf 1}^T \Ell - \lambda_2 {\bf 1}^T\hat{\bf d} \\  
     \label{eq:constraint1}
{\rm s.t.}\;\; & {\rm tr}({\bf A}_{ij}^k {\bf Y}^k) = \hat{d}_{ij}^k\;,\\
               & \begin{bmatrix} 1 & {\Ell^k}^T \\ {\Ell^k} & {\bf Y}^k \end{bmatrix} \succeq 0\;,\\
               & \hat{d}_{ij}^k \leq \hat{g}_{ij}\;,\; 
               \label{eq:formulation-gij2}
               {\bf 1}^T \hat{\bf g} = 1\;,\\
               & \ell_i^k \geq 0, \;\hat{d}_{ij}^k \geq 0\;,
               \forall \big(i,j\in\mathcal{N}(i),k\big)\;,
\end{align}
\end{subequations}
where $\lambda_1$, $\lambda_2$ are two positive trade-off parameters, and ${\bf 1}$ is an all-one column vector with appropriate dimensions. Note that we remove a term of $\hat{\bf g}$ in the objective~\eqref{eq:obj} because we have $\sum_{i,j,k} \hat{g}_{ij} = m$, which is a constant. Our formulation consists of a linear objective subject to linear constraints and SDP constraints, which is known as a convex problem. This convex SDP problem can be solved effectively by any modern SDP solver to a global optimum. 

\paragraph{Incomplete Data.}
Incomplete measurements are quite common due to occlusions. To handle incomplete measurements, we can introduce a set of visibility masks ${\bf{W}} \doteq \{w_i^k\}$, where $w_i^k = 1$ if the $i^{\rm th}$ point is visible in frame $k$, otherwise $w_i^k = 0$. With the visibility masks, the terms related to $\ell_i^k$ become $w_i^k\ell_i^k$ and the terms related to $d_{ij}^k$ become $w_i^kw_j^kd_{ij}^k$. The problem is still convex and solvable with any SDP solver. 
Here we assume that the number of visible points in one frame is greater than the neighborhood size; otherwise, we remove that frame.

\subsection{3D Reconstruction from Legs}
\label{sec:5.2}
Under the perspective camera model, we can relate ${\bf Q}_i^k$ and ${\bf q}_i^k$ with
\begin{equation}
{\bf Q}_i^k = z_i^k \hat{\bf q}_i^k
            = l_i^k \hat{\bf q}_i^k/\|\hat{\bf q}_i^k\|_2\;,
\end{equation}
where $\hat{\bf q}_i^k = {\bf K}^{-1}[{{\bf q}_i^k}^T \; 1]^T$. After we get the solutions for all legs $\ell_i^k$, we can then substitute them back to the above equation to compute the 3D coordinates for all points.

\paragraph{Degenerate Cases.} Our system becomes degenerate if there is only pure rotation (around the camera center) in the scene. In fact, pure rotation over the camera center do not change the angles between two vectors, \eg, ${\bf v}_1$ and ${\bf v}_2$, 
\begin{equation}
\cos(\theta) = \frac{{\bf v}_1^T{\bf v}_2}{\|{\bf v}_1\|\|{\bf v}_2\|} 
             = \frac{({\bf R}{\bf v}_1)^T({\bf R}{\bf v})_2}{\|{\bf R}{\bf v}_1\|\|{\bf R}{\bf v}_2\|}\;,
\end{equation}
where the equations hold because ${\bf R}^T{\bf R} = {\bf I}$ and rotation on vectors does not change their length. So if there is only pure rotation in the scene, our system will become under-constrained. This also corresponds to the fact in epipolar geometry that pure rotation cannot be  explained by the essential/fundamental matrix (but homography instead). Another degenerate case is when the camera model is close to orthographic. In this case, the viewing angles are all close to zero, which makes our formulation unsolvable.

\section{Experiments}

\begin{figure}[!t]
\centering
\includegraphics[width=0.95\linewidth]{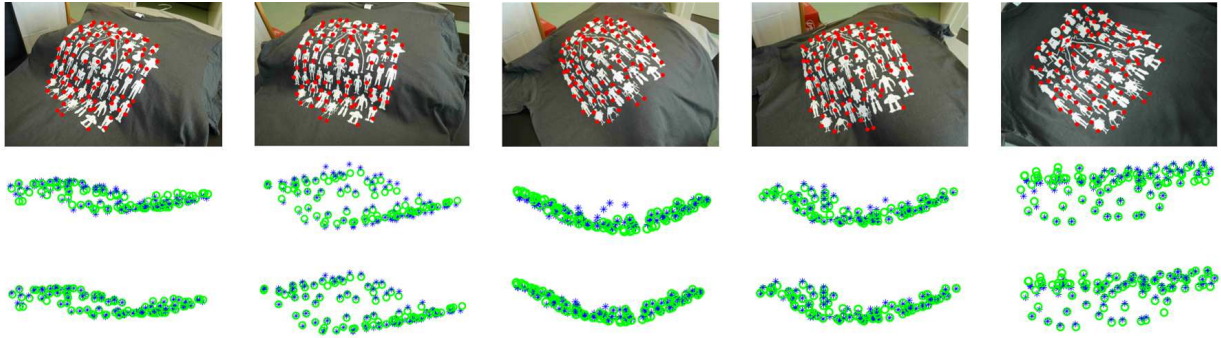}
\caption{Qualitative comparison of the 3D reconstruction results on the T-shirt dataset. The green circles plot the ground truth 3D points, and the blue stars show the reconstructed 3D points. {\bf Top row}: 2D images with feature points in red dots. {\bf Middle row}: results of~\cite{chhatkuli2016inextensible}. {\bf Bottom row}: results of our method. Best viewed on screen with zoom-in.}
\label{fig:tshirt}
\end{figure}

We compare our method with four baselines for rigid and non-rigid 3D shape reconstruction. These baselines include: the rigid ``structure-without-motion'' method for a perspective camera in~\cite{li2010multi}, the non-convex soft-inextensibility based NRSfM method for an orthographic camera in~\cite{Soft-Inextensibility-Template-Free:ECCV-2012}, the prior-free low-rank factorization based NRSfM method for an orthographic camera in~\cite{dai2014simple}, and the second-order cone programming based NRSfM method for a perspective camera~\cite{chhatkuli2016inextensible}. For the baselines, we use the source codes provided by the authors. We implement our method in Matlab and use the MOSEK~\cite{mosek} 
SDP solver to solve our formulation. We fix all the parameters of the baseline methods to the optimal values. We find that our method is not sensitive to the parameters $\lambda_1$ and $\lambda_2$, and set $\lambda_1 = 1$ and $\lambda_2 = 20$ for all our experiments, which are obtained by validating on a separate dataset. Due the limit of space, our qualitative reconstruction results on all synthetic datasets are provided in the supplementary videos.

The metrics we use to evaluate the performance are the 3D Root Mean Square Error (RMSE) (in mm) and the relative 3D error (denoted as R-Err) (in \%), which are respectively defined as
\begin{align*}
{\rm RMSE} &= \frac{1}{m}\sum_k\sqrt{\frac{1}{n}\sum_i\|\bar{\bf Q}_i^k - {\bf Q}_i^k\|^2_2}\;, \\
{\rm R\mbox{-}Err} &= \frac{1}{m}\sum_k\frac{\|\bar{\bf Q}^k - {\bf Q}^k \|_F}{\|\bar{\bf Q}^k\|_F}\times 100\%\;,
\end{align*}
where $\bar{\bf Q}_i^k$ is the ground truth coordinates of point $i$ in frame $k$. We always have a scale ambiguity for all structure-from-motion methods. For methods that use a perspective camera model, we re-scale their reconstructions to best align them with the ground truth before computing the errors. For methods that use an orthographic camera model, we do Procrustes analysis to solve for a similarity transformation that best aligns the reconstructions with the ground truth.

\subsection{Non-rigid Structure from Motion}

Our method and~\cite{chhatkuli2016inextensible} rely on constructing a K-NNG. For both methods, we use the same K-NNG and fix the neighborhood size $K$ as 20 for this set of experiments.

\vspace{-0.3cm}
\paragraph{The Flag (Semi-Synthetic) Dataset.} This flag dataset \cite{white2007capturing} consists of an image sequence of a fabric flag waving in the wind. 
The ground truth 3D points are provided in this dataset, but neither 2D projection trajectories nor camera calibrations are available. We subsample the 3D points in each frame and generate the input data  from a virtual perspective camera with the field-of-view angle as $81.69^{\circ}$. The final sequence contains 90 points (on each frame) and 50 frames. We report the 3D RMSE and mean relative 3D error in Table~\ref{tab:flag}. Note that our method achieves the lowest 3D reconstruction error among all the competing methods.

\begin{table}[h]
\centering
\caption{Mean 3D errors for the Flag Paper dataset.}
\label{tab:flag}
\begin{tabular}{ l c c c  c}
         & \cite{Soft-Inextensibility-Template-Free:ECCV-2012}  & \cite{dai2014simple}    & \cite{chhatkuli2016inextensible} & Ours \\
            \hline
  RMSE   & 41.92 &  26.23    &  21.08                            &  {\bf 16.75} \\        
  R-Err  & 12.76\% &  7.51\%   &  6.38\%                          &  {\bf 5.07\%} \\
 \hline
\end{tabular}
\vspace{-0.3cm}
\end{table}

\vspace{-0.3cm}
\paragraph{The KINECT Paper, Hulk, and T-Shirt Datasets.} 
The KINECT paper dataset~\cite{varol2012constrained} contains an image sequence of smoothly deforming well-textured paper captured by a KINECT camera.  The camera calibration and ground truth 3D are provided. We use the trajectories provided by~\cite{chhatkuli2016inextensible}, which was obtained by tracking interest points in this sequence using a flow-based method of~\cite{garg2013dense}. The trajectories are complete, semi-dense and outlier-free. Due to the large number of points and frames, we subsample the points and frames in this dataset and get a sequence with 151 points (on each frame) and 23 frames. 

The Hulk dataset~\cite{chhatkuli2014non} consists of 21 images taken at different unrelated smooth deformations. The deforming scene is a well-textured paper cover of a comics. The intrinsic camera calibration matrix, 3D ground truth shape and 2D feature trajectories are provided in this dataset. This dataset contains 122 trajectories in 21 views.

The T-Shirt dataset~\cite{chhatkuli2014non} consists of 10 images taken for a deforming T-shirt. As in the Hulk dataset, the intrinsic camera calibration matrix, 3D ground truth shape and 2D feature trajectories are all provided in this dataset. This dataset contains 85 point trajectories in 10 frames.

\begin{figure}[!htp]
\centering
\includegraphics[width=1.00\linewidth]{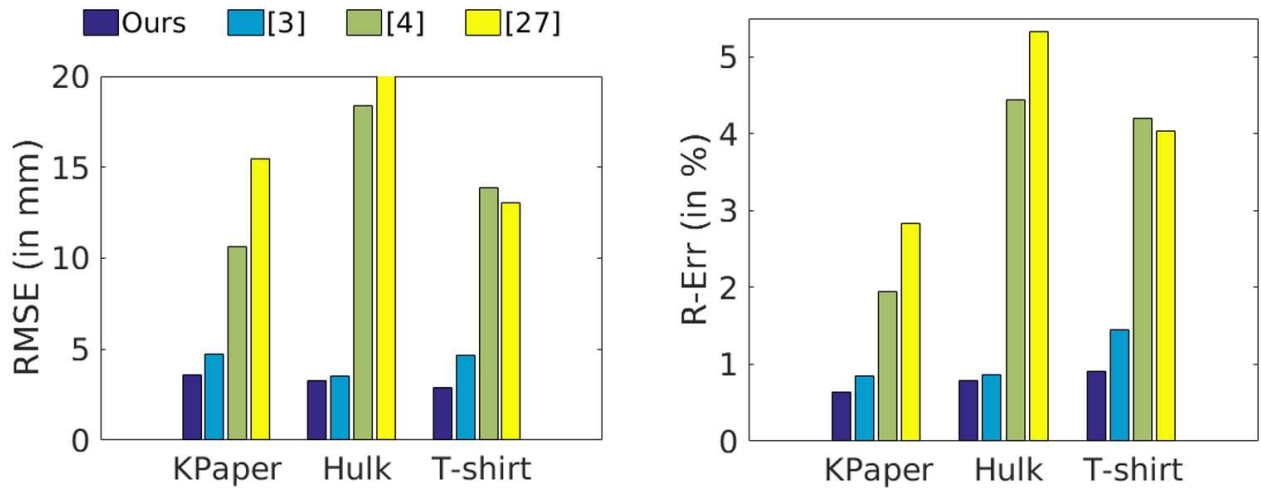}
\caption{Mean 3D reconstruction errors for the KINECT Paper (denoted as ``KPaper'' in the figure), Hulk, and T-shirt Datasets. Our method (plotted in blue) achieves the lowest 3D reconstruction errors on all three datasets.}
\label{fig:kpaperhulktshirt}
\vspace{-0.3cm}
\end{figure}

We show the mean 3D errors of our method and the baselines in Figure~\ref{fig:kpaperhulktshirt}. We can see that our method achieves the lowest 3D reconstruction error on all the three datasets. We also give a qualitative comparison with the best-performing baseline~\cite{chhatkuli2016inextensible} in Figure~\ref{fig:tshirt}.

\vspace{-0.3cm}
\paragraph{The Jumping Trousers Dataset with missing data.} This dataset~\cite{white2007capturing} contains 3D ground truth points for jumping trousers obtained from cloth motion capture. The complete 2D trajectories are generated  by projecting the 3D points through a virtual perspective camera. However, due to self-occlusions, the 2D trajectories would have a considerable amount of missing entries, and the visibility masks are provided in the original data. We subsample the points and frames, and get a sequence of 97 points and 29 frames. Since the first two baselines \cite{Soft-Inextensibility-Template-Free:ECCV-2012,dai2014simple} cannot handle incomplete data, we input complete trajectories for them. We use the incomplete trajectories for \cite{chhatkuli2016inextensible} and our method as the two methods can handle incomplete data.\footnote{Note, for incomplete data, we only compute average 3D reconstruction error for visible points. And also note that this comparison is unfair for \cite{chhatkuli2016inextensible} and our method as the other two use complete data.} The results are reported in Table~\ref{tab:trousers}. Our method, with incomplete data as input, outperforms all the other baselines. 
\begin{table}[h]
\centering
\caption{Mean 3D errors for the Jumping Trousers dataset.}
\label{tab:trousers}
\begin{tabular}{ l c c c  c}

            & \cite{Soft-Inextensibility-Template-Free:ECCV-2012} & \cite{dai2014simple} & \cite{chhatkuli2016inextensible} & Ours \\
            \hline
  RMSE      &  190.17   &  49.97     &  44.05                           &  {\bf 37.70}  \\        
  R-Err     &  55.10\%  &   12.67\%  &  13.57\%                         &  {\bf 11.65\%}  \\
 \hline
\end{tabular}
\vspace{-0.1cm}
\end{table}

From this set of experiments, we have shown that our method consistently outperforms all the baselines. We note that on those datasets there is always a significant performance gap between those orthographic camera model based methods (\cite{Soft-Inextensibility-Template-Free:ECCV-2012,dai2014simple}) and those perspective camera model based methods (\cite{chhatkuli2016inextensible} and ours). In the following experiments, we will only compare with the perspective camera model based methods (\cite{li2010multi,chhatkuli2016inextensible}).

\vspace{-0.3cm}
\paragraph{Robustness to various numbers of points/views, different levels of missing data and noise.}

\begin{figure}[!t]
\centering
\includegraphics[width=1.0\linewidth]{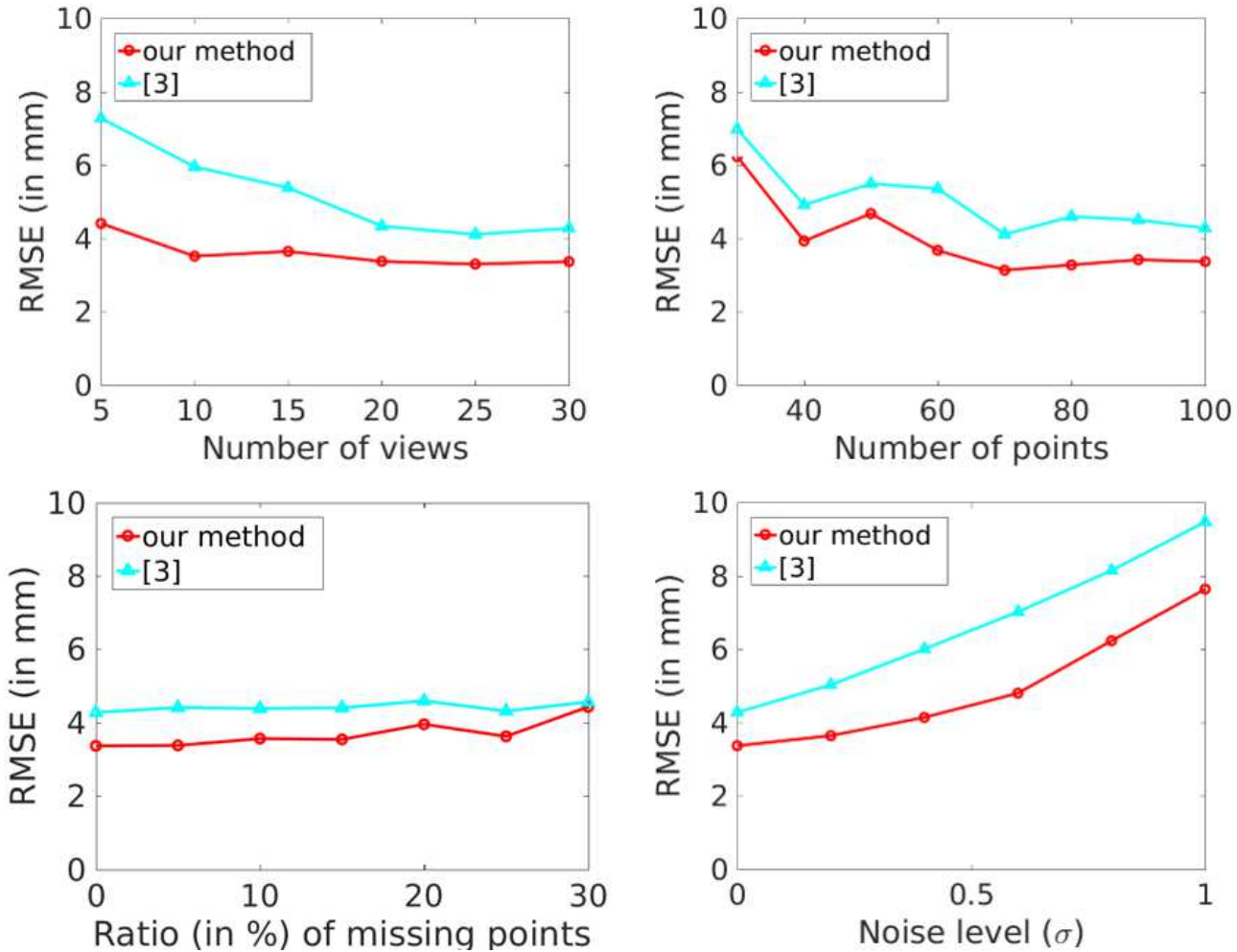}
\caption{Performance evaluation on the KINECT Paper dataset with increasing number of points/views, increasing ratios of missing data, and increasing levels of synthetic Gaussian noise.}
\label{fig1:kpaper_synthetic}
\vspace{-0.2cm}
\end{figure}

In Figure~\ref{fig1:kpaper_synthetic}, we show the performance of our method and the best-performing baseline \cite{chhatkuli2016inextensible} on the KINECT paper dataset with increasing number of points/views, increasing ratios of missing data, and increasing levels of synthetic zero-mean Gaussian noise (with various standard deviations $\sigma$). The default experimental setting is with 100 points and 30 views, and the parameters are fixed as $\lambda_1 = 1$, $\lambda_2 = 20$, and $K = 20$. We can see that our method consistently outperforms the baseline method in all scenarios, which verifies the robustness of our method. We believe that our superior performance comes from the novel maximizing rigidity regularization, which better explains the image observations.

\subsection{Rigid Structure from Motion}

In this set of experiments, we test our method for rigid structure reconstruction. Since our method does not utilize the rigidity prior of the scene, we can well expect that our method performs worse than the specifically designed rigid method. The main goal of this set of experiments is thus to show that our method can achieve comparable rigid structure reconstruction to the rigid method. We compare our method with the best-performing baseline~\cite{chhatkuli2016inextensible} for non-rigid structure from motion, and another method~\cite{li2010multi} specifically designed for rigid structure from motion. The neighborhood size is set as 20 for all methods.

\begin{figure}[!t]
\centering
\begin{tabular}{cc}
\hspace{-0.2cm}\includegraphics[width=0.53\linewidth]{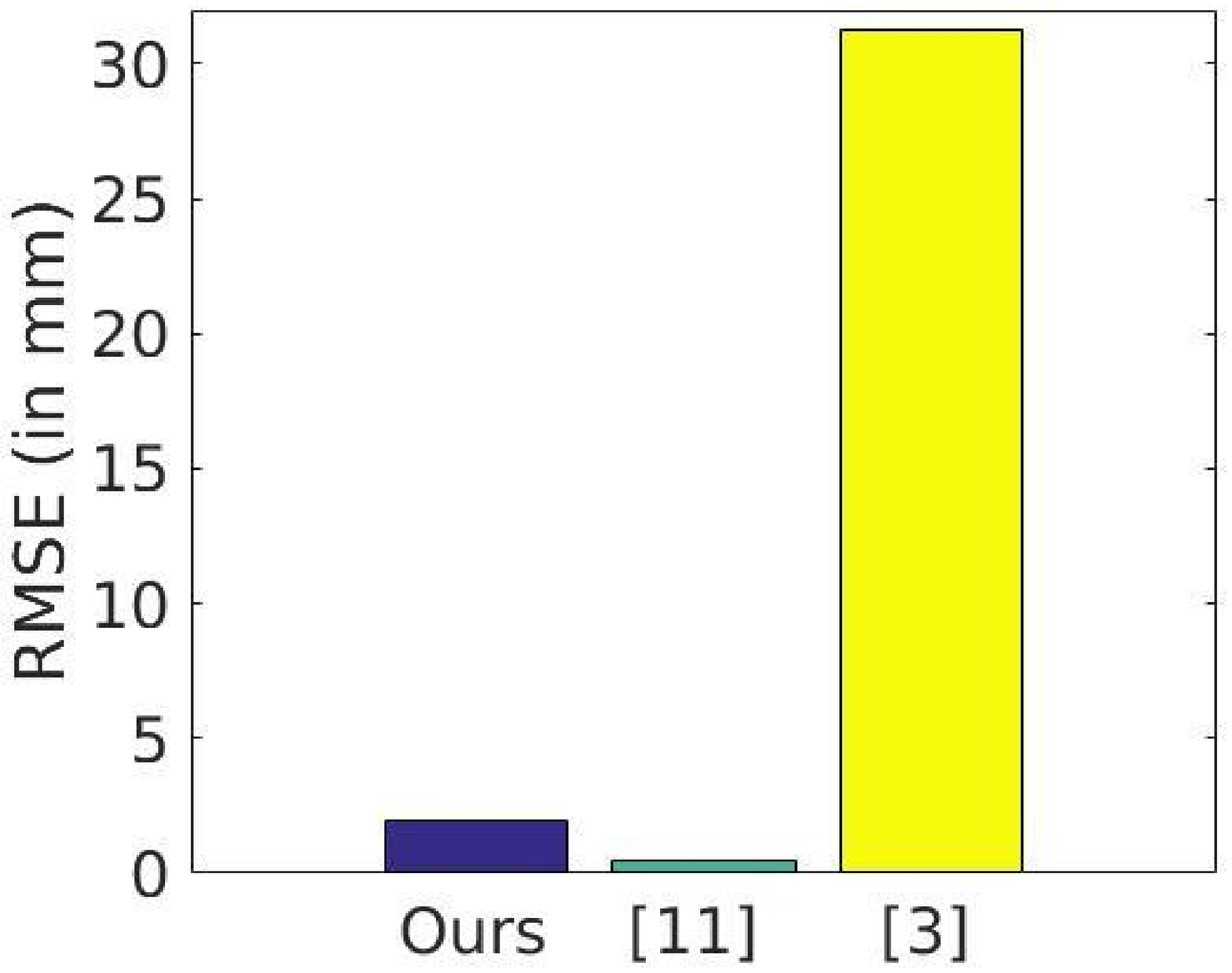} &\hspace{-0.7cm}
\includegraphics[width=0.53\linewidth]{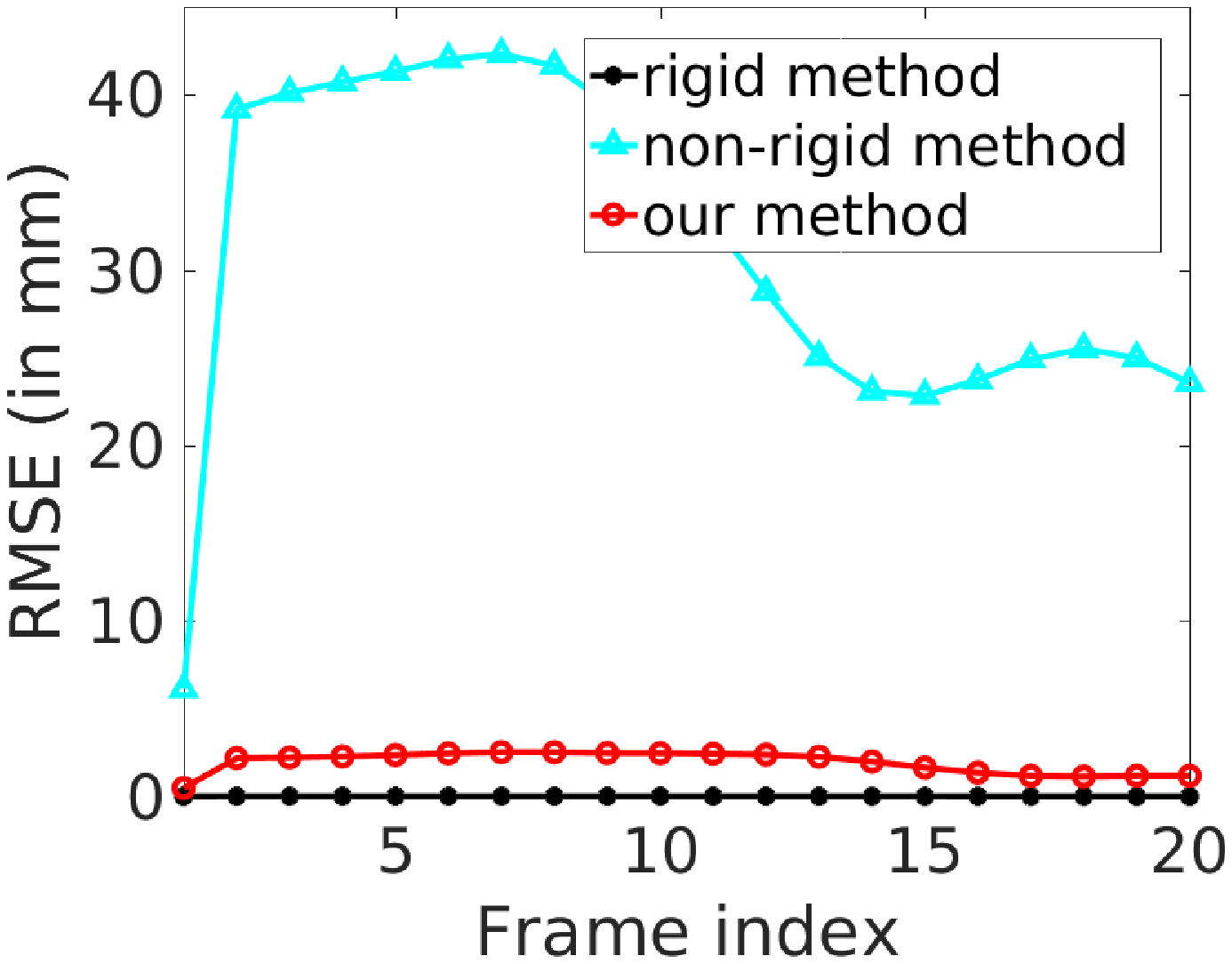}
\end{tabular}
\caption{{\bf Left:} mean 3D errors for the synthetic rigid dataset. {\bf Right:} the RMSE (in mm) for each frame of the synthetic sequence by the rigid method~\cite{li2010multi} (in black dots), the non-rigid method~\cite{chhatkuli2016inextensible} (in cyan dots), and our method (in red dots).}
\label{fig:rigid}
\vspace{-0.2cm}
\end{figure}

\vspace{-0.3cm}
\paragraph{Rigid Synthetic Dataset.} We verify our method for rigid structure computation on a synthetic dataset. To generate the data, we subsample the ground truth 3D points of one frame of the KINECT paper dataset~\cite{varol2012constrained}, and apply a transformation (rotation and translation) to these points over time. After a perspective projection, we get a sequence for rigid motion with 61 points and 20 frames. The mean 3D reconstruction errors for all competing methods are reported in Figure~\ref{fig:rigid}. We also plot the RMSE (in mm) for each frame of the sequence in Figure~\ref{fig:rigid} and compare our method with the state-of-the-art non-rigid SfM method~\cite{chhatkuli2016inextensible} and the rigid SfM method~\cite{li2010multi}. It's no surprising that \cite{li2010multi} achieves the lowest reconstruction errors in this rigid dataset as it utilizes the prior knowledge that the scene is rigid. Our method, without inputting any prior knowledge of the scene rigidity, gets close results to~\cite{li2010multi} and significantly outperforms the NRSfM method~\cite{chhatkuli2016inextensible}.


\vspace{-0.3cm}
\paragraph{The Model House Dataset.} We use the VGG model house dataset~\footnote{\url{http://www.robots.ox.ac.uk/~vgg/data/data-mview.html}}
as the real-world dataset for rigid SfM. The camera projection matrices, 2D feature coordinates and 3D ground truth points are provided in this dataset, and the 2D measurements contain moderate amount of noise. The camera intrinsic matrices are computed from camera projection matrices using R-Q decomposition~\cite{hartley2003multiple}. We generate a sequence with complete feature point trajectories of 95 points and 7 frames. We report the 3D reconstruction errors of all methods in Table~\ref{tab:house}. Again, our method obtains close results to~\cite{li2010multi} and lower reconstruction error than the NRSfM method~\cite{chhatkuli2016inextensible}.
\begin{table}[h]
\centering
\caption{Mean 3D errors for the Model House dataset.}
\label{tab:house}
\begin{tabular}{ l c c  c}

           & \cite{li2010multi}   & \cite{chhatkuli2016inextensible} & Ours \\
            \hline
  RMSE     & 0.158    &  0.200                           &   0.162 \\        
  R-Err    & 2.95\%   &  3.73\%                          &   3.02\%  \\
 \hline
\end{tabular}
\end{table}

\subsection{Articulated Motion Reconstruction}

In this set of experiments, we evaluate our method for the 3D reconstruction of articulated motions, and compare our method with the best-performing baseline~\cite{chhatkuli2016inextensible}.

\vspace{-0.3cm}
\paragraph{Synthetic Articulated Dataset.} We first test our method on two synthetic sequences where the objects undergo articulated motions. To generate the synthetic data, we take a subset of the ground truth 3D points in the first image of the KINECT paper dataset~\cite{varol2012constrained} and divide them into two groups. We synthesize two kinds of articulated motions: (i) the point-articulated motion (denoted as ``point-articulated'' in Table~\ref{tab:articulated-synthetic}), \ie, the two groups of points rotate around a common point in the dataset and meanwhile undertake the same translations through time; (ii) the axis-articulated motion (denoted as ``axis-articulated'' in Table~\ref{tab:articulated-synthetic}), \ie, the two groups of points rotate around a common axis in the dataset and also undertake the same translations. The 2D feature points are generated by projecting these 3D points with a virtual perspective camera. We finally get two synthetic sequences with 61 points and 19 frames. We report the RMSE (in millimeter) and the mean relative 3D error in Table~\ref{tab:articulated-synthetic}. Our method achieves much lower 3D reconstruction error than the baseline method~\cite{chhatkuli2016inextensible}.

\begin{table}[h]
\centering
\caption{RMSE (in mm) and mean relative 3D error (shown in brackets) in percentage (\%) for the synthetic articulated data.}
\label{tab:articulated-synthetic}
\begin{tabular}{ c  c  c}

     sequence         & \cite{chhatkuli2016inextensible} & Ours \\
            \hline
  point-articulated      &   17.48 (2.45\%)             &  {\bf 7.70 (1.11\%)} \\      
  axis-articulated       &   9.13 (1.36\%)              &  {\bf 3.07 (0.45\%)} \\ 
 \hline
\end{tabular}
\vspace{-0.2cm}
\end{table}

\begin{figure}[!t]
\centering
\begin{tabular}{cc}
\hspace{-0.4cm}\includegraphics[width=0.55\linewidth]{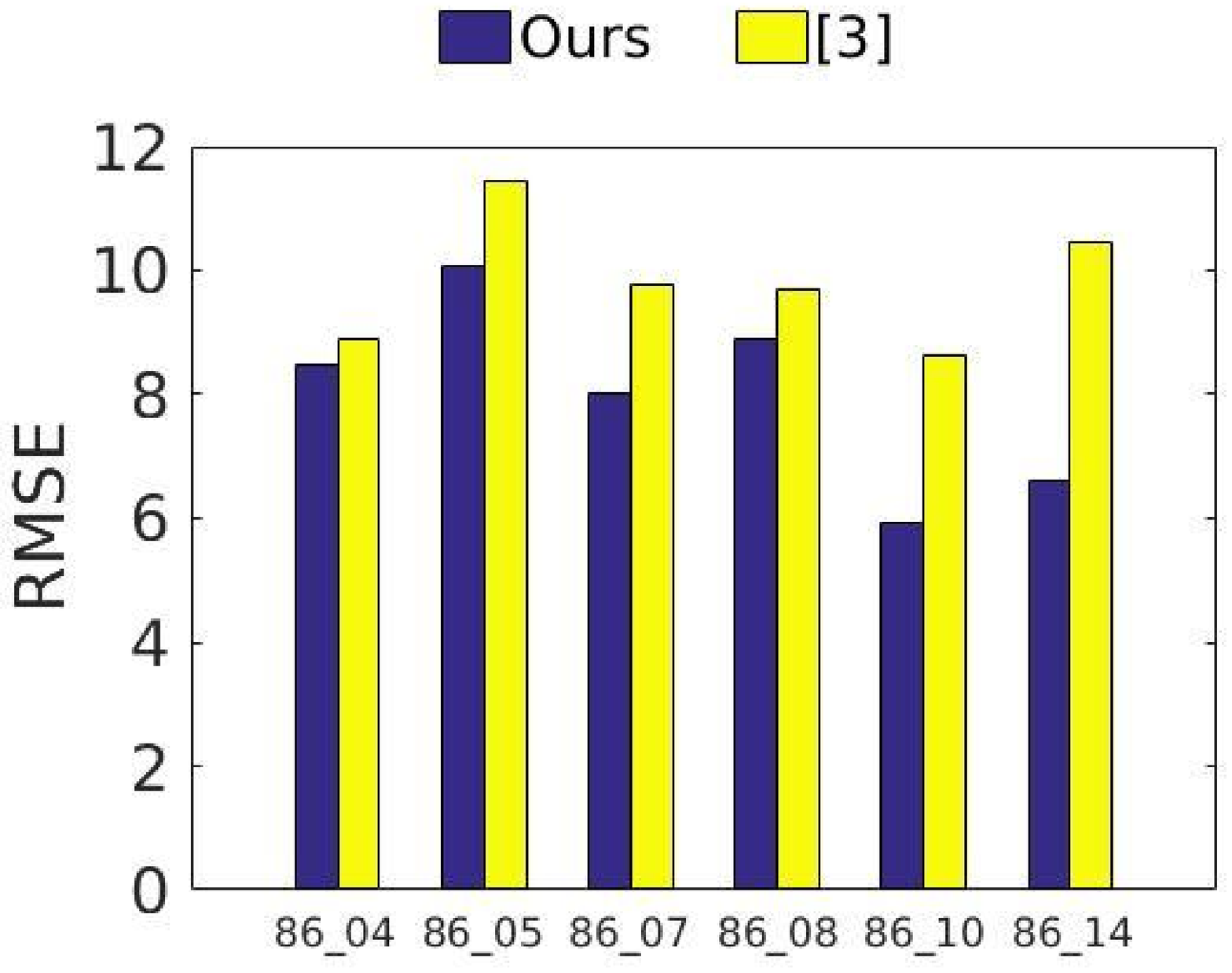} & 
\hspace{-0.6cm}\includegraphics[width=0.55\linewidth]{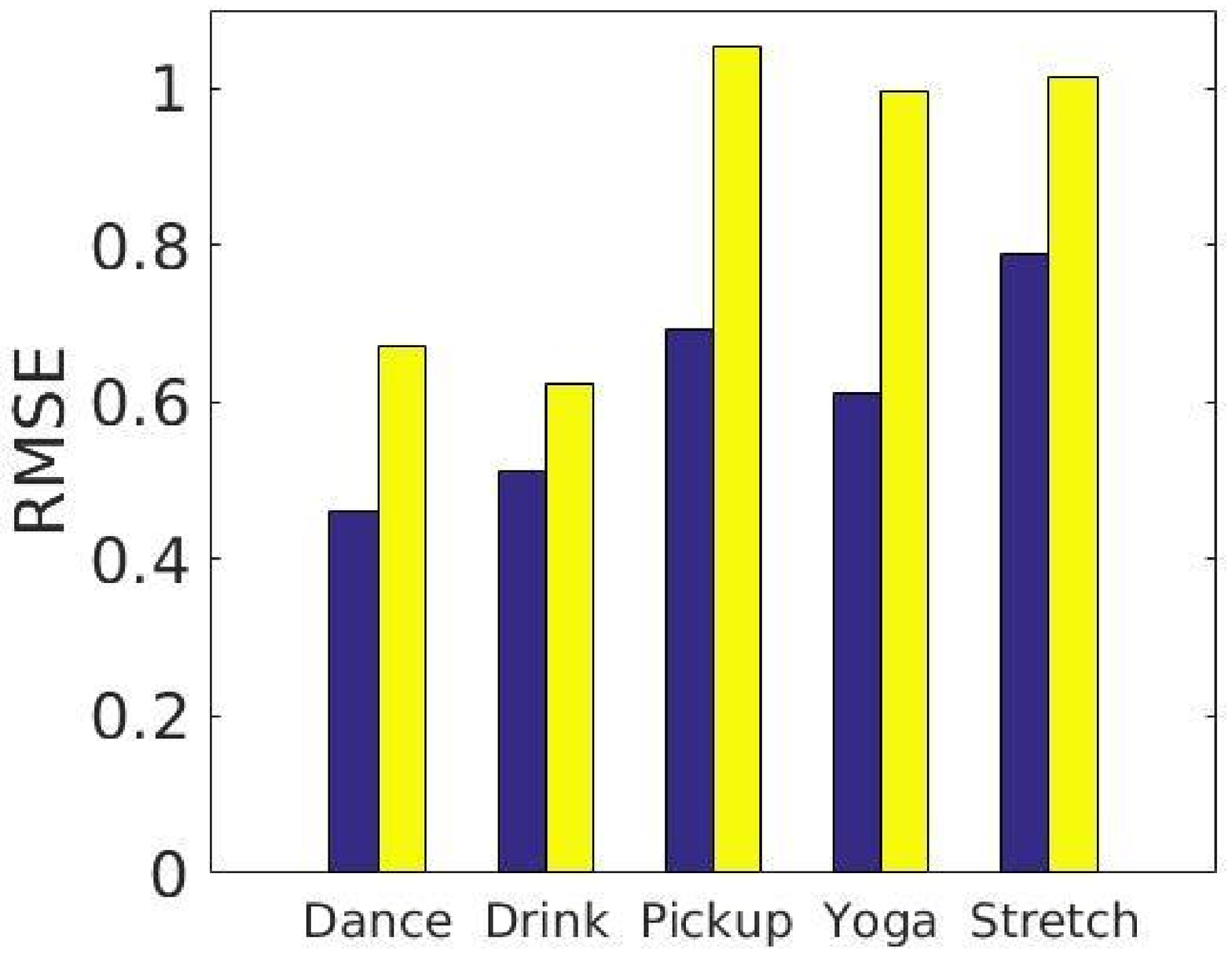}
\end{tabular}
\caption{{\bf Left}: mean 3D reconstruction errors on six sampled sequences (86\_04, 86\_05, 86\_07, 86\_08, 86\_10, and 86\_14) of CMU Mocap Database. {\bf Right}: mean 3D reconstruction errors on Dance, Drink, Pickup, Yoga, and Stretch sequences.}
\label{fig:rmse_mocap}
\vspace{-0.2cm}
\end{figure}

\begin{figure}[!t]
\centering
\includegraphics[width=0.90\linewidth]{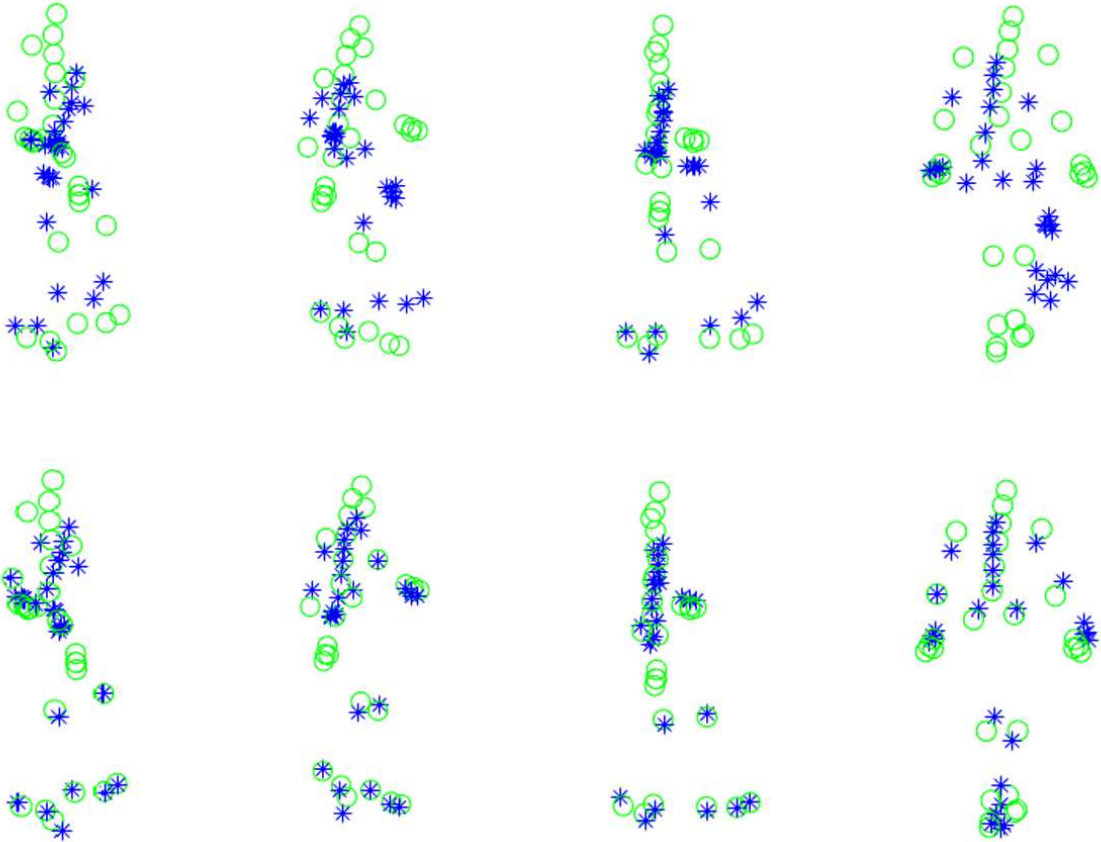}
\caption{Qualitative comparison of the 3D reconstruction results on the CMU Mocap Database. The green circles plot the ground truth 3D points, and the blue stars show the reconstructed 3D points. {\bf Top row}: results of~\cite{chhatkuli2016inextensible}. {\bf Bottom row}: results of our method.}
\label{fig:mocap}
\vspace{-0.2cm}
\end{figure}

\vspace{-0.3cm}
\paragraph{Human Motion Capture Database.} We sample six sequences in the CMU Mocap Database~\footnote{\url{http://mocap.cs.cmu.edu/}} and five sequences (Dance, Drink, Pickup, Yoga, and Stretch sequences) used in~\cite{dai2014simple} to form the human motion capture database. For the latter five sequences, the data are centered to fit the factorization-based methods, so we further add random translations to each frame. Each sequence of this database consists of 28 (for CMU Mocap), 41 (for Drink, Pickup, Yoga, Stretch) or 75 (for Dance) points with 3D ground truth coordinates. The input data are generated from a virtual camera with perspective projection. We uniformly sub-sample the frames of each sequence with a sample rate 10 (\ie, $1:10:{\rm end}$) for CMU Mocap and a sample rate 5 for other sequences, producing sequences with 52 to 335 frames. For CMU Mocap, we set the neighborhood size $K$ as 28 for all competing methods, which lets us to use all available points to build the edges; for other sequences, we set $K$ as 20. We show the quantitative results of our method and the baseline method in Figure~\ref{fig:rmse_mocap}, and also give a qualitative comparison of the 3D reconstruction results on this dataset in Figure~\ref{fig:mocap}. We can see that our method consistently outperforms the baseline~\cite{chhatkuli2016inextensible}.

\section{Concluding Remarks}

In this paper, we have revisited Ullman's principle of maximizing rigidity and proposed a novel convex rigidity measure that can be incorporated into a modern structure reconstruction framework to unify both rigid and non-rigid SfM from multiple perspective images. Our reconstruction method relies on directly building viewing triangles, thus not requiring to estimate camera poses. Importantly, our formulation (after SDP relaxations) is convex such that a global optimal solution is guaranteed. We have verified the efficacy of our method by extensive experiments on multiple rigid, non-rigid and articulated datasets.

\vspace{-0.3cm}
\paragraph{Limitation and Future Work.} The computational bottleneck of our method lies in solving the SDPs. For a sequence of $m$ views and $n$ points (for each view), we need to solve $m$ SDPs of size $(n+1)\times (n+1)$. Using an interior-point method, one SDP has a worst-case complexity of $O(n^{4.5}{\rm log}(1/\epsilon))$ given a solution accuracy $\epsilon>0$~\cite{luo2010semidefinite}, which remains the limiting factor preventing us from testing on modern large-scale datasets. In the future, we aim to explore the possibility of applying modern large-scale SDP solver, such as~\cite{zhao2010newton,yang2015sdpnal+}, to solve our problem more efficiently. Furthermore, we also plan to investigate how to address the degenerate cases as discussed in Sec.~\ref{sec:5.2}. 
{\small
\bibliographystyle{ieee}
\bibliography{SFM_Reference}

\begin{thebibliography}{10}\itemsep=-1pt

\bibitem{boyd2004convex}
S.~Boyd and L.~Vandenberghe.
\newblock {\em Convex optimization}.
\newblock Cambridge university press, 2004.

\bibitem{chhatkuli2014non}
A.~Chhatkuli, D.~Pizarro, and A.~Bartoli.
\newblock Non-rigid shape-from-motion for isometric surfaces using
  infinitesimal planarity.
\newblock In {\em BMVC}, 2014.

\bibitem{chhatkuli2016inextensible}
A.~Chhatkuli, D.~Pizarro, T.~Collins, and A.~Bartoli.
\newblock Inextensible non-rigid shape-from-motion by second-order cone
  programming.
\newblock In {\em Proceedings of the IEEE Conference on Computer Vision and
  Pattern Recognition}, pages 1719--1727, 2016.

\bibitem{dai2014simple}
Y.~Dai, H.~Li, and M.~He.
\newblock A simple prior-free method for non-rigid structure-from-motion
  factorization.
\newblock {\em International Journal of Computer Vision}, 107(2):101--122,
  2014.

\bibitem{d2003relaxations}
A.~d’Aspremont and S.~Boyd.
\newblock Relaxations and randomized methods for nonconvex {QCQP}s.
\newblock {\em EE392o Class Notes, Stanford University}, 2003.

\bibitem{garg2013dense}
R.~Garg, A.~Roussos, and L.~Agapito.
\newblock Dense variational reconstruction of non-rigid surfaces from monocular
  video.
\newblock In {\em Proceedings of the IEEE Conference on Computer Vision and
  Pattern Recognition}, pages 1272--1279, 2013.

\bibitem{grzywacz1987incremental}
N.~M. Grzywacz and E.~C. Hildreth.
\newblock Incremental rigidity scheme for recovering structure from motion:
  Position-based versus velocity-based formulations.
\newblock {\em JOSA A}, 4(3):503--518, 1987.

\bibitem{hartley2003multiple}
R.~Hartley and A.~Zisserman.
\newblock {\em Multiple view geometry in computer vision}.
\newblock Cambridge university press, 2003.

\bibitem{hartley1997defense}
R.~I. Hartley.
\newblock In defense of the eight-point algorithm.
\newblock {\em IEEE Transactions on pattern analysis and machine intelligence},
  19(6):580--593, 1997.

\bibitem{hartley1997triangulation}
R.~I. Hartley and P.~Sturm.
\newblock Triangulation.
\newblock {\em Computer vision and image understanding}, 68(2):146--157, 1997.

\bibitem{li2010multi}
H.~Li.
\newblock Multi-view structure computation without explicitly estimating
  motion.
\newblock In {\em Computer Vision and Pattern Recognition (CVPR), 2010 IEEE
  Conference on}, pages 2777--2784. IEEE, 2010.

\bibitem{longuet1987computer}
H.~C. Longuet-Higgins.
\newblock A computer algorithm for reconstructing a scene from two projections.
\newblock {\em Readings in Computer Vision: Issues, Problems, Principles, and
  Paradigms, MA Fischler and O. Firschein, eds}, pages 61--62, 1987.

\bibitem{luo2010semidefinite}
Z.-Q. Luo, W.-k. Ma, A.~M.-C. So, Y.~Ye, and S.~Zhang.
\newblock Semidefinite relaxation of quadratic optimization problems.
\newblock {\em IEEE Signal Processing Magazine}, 27(3):20, 2010.

\bibitem{mosek}
A.~Mosek.
\newblock The {MOSEK} optimization software.
\newblock {\em Online at http://www. mosek. com}, 54:2--1, 2010.

\bibitem{parashar2015rigid}
S.~Parashar, D.~Pizarro, A.~Bartoli, and T.~Collins.
\newblock As-rigid-as-possible volumetric shape-from-template.
\newblock In {\em Proceedings of the IEEE International Conference on Computer
  Vision}, pages 891--899, 2015.

\bibitem{pardalos1991quadratic}
P.~M. Pardalos and S.~A. Vavasis.
\newblock Quadratic programming with one negative eigenvalue is {NP}-hard.
\newblock {\em Journal of Global Optimization}, 1(1):15--22, 1991.

\bibitem{Perriollat:BMVC08}
M.~Perriollat, R.~Hartley, and A.~Bartoli.
\newblock Monocular template-based reconstruction of inextensible surfaces.
\newblock In {\em British Machine Vision Conference}, 2008.

\bibitem{Perriollat_IJCV_2010}
M.~Perriollat, R.~Hartley, and A.~Bartoli.
\newblock Monocular template-based reconstruction of inextensible surfaces.
\newblock {\em International Journal of Computer Vision}, 2010.

\bibitem{sahni1974computationally}
S.~Sahni.
\newblock Computationally related problems.
\newblock {\em SIAM Journal on Computing}, 3(4):262--279, 1974.

\bibitem{salzmann2011linear}
M.~Salzmann and P.~Fua.
\newblock Linear local models for monocular reconstruction of deformable
  surfaces.
\newblock {\em IEEE Transactions on Pattern Analysis and Machine Intelligence},
  33(5):931--944, 2011.

\bibitem{sorkine2007rigid}
O.~Sorkine and M.~Alexa.
\newblock As-rigid-as-possible surface modeling.
\newblock In {\em Symposium on Geometry processing}, volume~4, 2007.

\bibitem{sturm1996factorization}
P.~Sturm and B.~Triggs.
\newblock A factorization based algorithm for multi-image projective structure
  and motion.
\newblock In {\em European conference on computer vision}, pages 709--720,
  1996.

\bibitem{Locally-Rigid:CVPR-2010}
J.~Taylor, A.~D. Jepson, and K.~N. Kutulakos.
\newblock Non-rigid structure from locally-rigid motion.
\newblock In {\em Computer Vision and Pattern Recognition (CVPR), 2010 IEEE
  Conference on}, pages 2761--2768, June 2010.

\bibitem{tomasi1992shape}
C.~Tomasi and T.~Kanade.
\newblock Shape and motion from image streams under orthography: a
  factorization method.
\newblock {\em International Journal of Computer Vision}, 9(2):137--154, 1992.

\bibitem{ullman1979interpretation}
S.~Ullman.
\newblock The interpretation of structure from motion.
\newblock {\em Proceedings of the Royal Society of London B: Biological
  Sciences}, 203(1153):405--426, 1979.

\bibitem{ullman1984maximizing}
S.~Ullman.
\newblock Maximizing rigidity: The incremental recovery of 3-d structure from
  rigid and nonrigid motion.
\newblock {\em Perception}, 13(3):255--274, 1984.

\bibitem{varol2012constrained}
A.~Varol, M.~Salzmann, P.~Fua, and R.~Urtasun.
\newblock A constrained latent variable model.
\newblock In {\em Computer Vision and Pattern Recognition (CVPR), 2012 IEEE
  Conference on}, pages 2248--2255, 2012.

\bibitem{Salzmann-Template-Free:ICCV-2009}
A.~Varol, M.~Salzmann, E.~Tola, and P.~Fua.
\newblock Template-free monocular reconstruction of deformable surfaces.
\newblock In {\em 2009 IEEE 12th International Conference on Computer Vision},
  pages 1811--1818, Sept 2009.

\bibitem{Soft-Inextensibility-Template-Free:ECCV-2012}
S.~Vicente and L.~Agapito.
\newblock Soft inextensibility constraints for template-free non-rigid
  reconstruction.
\newblock In {\em European Conference on Computer Vision}, pages 426--440,
  2012.

\bibitem{wang2016template}
X.~Wang, M.~Salzmann, F.~Wang, and J.~Zhao.
\newblock Template-free 3d reconstruction of poorly-textured nonrigid surfaces.
\newblock In {\em European Conference on Computer Vision}, pages 648--663.
  Springer, 2016.

\bibitem{white2007capturing}
R.~White, K.~Crane, and D.~A. Forsyth.
\newblock Capturing and animating occluded cloth.
\newblock In {\em ACM Transactions on Graphics (TOG)}, volume~26, page~34. ACM,
  2007.

\bibitem{yang2015sdpnal+}
L.~Yang, D.~Sun, and K.-C. Toh.
\newblock {SDPNAL}+: a majorized semismooth newton-cg augmented lagrangian
  method for semidefinite programming with nonnegative constraints.
\newblock {\em Mathematical Programming Computation}, 7(3):331--366, 2015.

\bibitem{zhao2010newton}
X.-Y. Zhao, D.~Sun, and K.-C. Toh.
\newblock A newton-cg augmented lagrangian method for semidefinite programming.
\newblock {\em SIAM Journal on Optimization}, 20(4):1737--1765, 2010.

\end{thebibliography}
}
\end{document}